\documentclass[10pt,twocolumn,letterpaper]{article}

\usepackage[pagenumbers]{cvpr} 

\usepackage{graphicx}
\usepackage{amsmath}
\usepackage{amssymb}
\usepackage{booktabs}
\usepackage[ruled, linesnumbered]{algorithm2e}
\usepackage{balance}
\usepackage[accsupp]{axessibility}

\usepackage[pagebackref,breaklinks,colorlinks]{hyperref}

\usepackage[capitalize]{cleveref}
\crefname{section}{Sec.}{Secs.}
\Crefname{section}{Section}{Sections}
\Crefname{table}{Table}{Tables}
\crefname{table}{Tab.}{Tabs.}

\def\cvprPaperID{7} 
\def\confName{CVPR}
\def\confYear{2022}

\begin{document}

\title{Adaptive Differential Filters for \\ Fast and Communication-Efficient Federated Learning}

\author{Daniel Becking \qquad Heiner Kirchhoffer \qquad Gerhard Tech\qquad Paul Haase \qquad
Karsten Müller \qquad \\ Heiko Schwarz \qquad Wojciech Samek\\
Fraunhofer Heinrich Hertz Institute (HHI), Berlin\\
{\tt\small \{firstname.lastname\}@hhi.fraunhofer.de}
}
\maketitle

\begin{abstract}
Federated learning (FL) scenarios inherently generate a large communication overhead by frequently transmitting neural network updates between clients and server. To minimize the communication cost, introducing sparsity in conjunction with differential updates is a commonly used technique. However, sparse model updates can slow down convergence speed or unintentionally skip certain update aspects, e.g., learned features, if error accumulation is not properly addressed. 
In this work, we propose a new scaling method operating at the granularity of convolutional filters which 1) compensates for highly sparse updates in FL processes, 2) adapts the local models to new data domains by enhancing some features in the filter space while diminishing others and 3) motivates extra sparsity in updates and thus achieves higher compression ratios, i.e., savings in the overall data transfer.
Compared to unscaled updates and previous work, experimental results on different computer vision tasks (Pascal VOC, CIFAR10, Chest X-Ray) and neural networks (ResNets, MobileNets, VGGs) in uni-, bidirectional and partial update FL settings show that the proposed method improves the performance of the central server model while converging faster and reducing the total amount of transmitted data by up to $377\times$.
\end{abstract}

\section{Introduction}
\label{sec:intro}

With current trends in ``\textit{Artificial Intelligence of Things}'' and the rapidly growing number of
intelligent devices~\cite{survey_edge_iot}, distributed computing on the edge becomes increasingly important. 
Federated learning (FL) allows multiple such edge devices to jointly train a deep learning model on their own local data and thus provides a basic level of privacy to the participating instances, since the local training data never leaves the devices. However, communication overhead is a major bottleneck hindering a straight-forward scalability of distributed learning systems. Client model updates or gradients are of the same size as the full model, which can be in the range of hundreds of megabytes for modern computer vision models~\cite{deeplabv3plus}. To enable distributed training with many edge devices, extensive research on weight (update) compression has been conducted in recent years to improve computational and communication efficiency~\cite{survey_fl_edge}.
Communication efforts can be reduced through two basic concepts: 1) by reducing the communication frequency of weight updates, i.e., multiple local iterations of weight updates are performed before transmission; or 2) by compressing the data to be transmitted, e.g., by applying sparsification, quantization or encoding methods.
Often, those methods slow down the training process, as they increase the number of iterations required to achieve a converged state of the neural network (NN) parameters. 

Moreover, finding optimal communication protocol-specific hyperparameters (e.g., the number of communication rounds, the fraction of participating clients, or sparsification rates) becomes infeasible as the number of clients increases. Therefore, the authors of~\cite{survey_fl_systemlevel} suggest to develop easy-to-tune or auto-tuning algorithms for the instances that compose the federation, i.e., the client networks, instead of considering their network architectures, optimizers, and regularizations as already established and aiming only at the best accuracy of the central server model by optimizing the communication protocol. 
This is one of the motivations for our proposed filter scaling method, which is a minimally invasive intervention in the computational graphs of the neural networks, but has a significant impact on the training process.
In this paper, we propose an FL pipeline in which compressed communication data and low communication frequency do not compromise on convergence speed. With our studies we show that \begin{itemize}
\setlength{\itemsep}{0.0cm}
\item in various federated computer vision tasks, our proposed method quickly adapts the knowledge of clients' NNs from a given domain to new domains
\item equipping NN layers with additional trainable scaling factors accelerates and regulates the entire FL process
\item thanks to our sparsification, encoding and scaling technologies, the amount of transmitted data is minimal
\item our proposed method leads to  considerable  speed ups compared to prior work and improves communication efficiency.
\end{itemize}

First, we discuss the relation to previous works (Section~\ref{sec:prior}), then we introduce our compression pipeline (Section~\ref{sec:compression}), the filter scaling mechanism and the training protocol (Section~\ref{sec:method}). Section~\ref{sec:exp} presents experimental results, including the review of different optimization methods, the effects of filter scaling, computational overheads, variation in the number of clients and comparison with prior work. Section~\ref{sec:conc} concludes the paper. 

\section{Relation to Prior Work}
\label{sec:prior}

For data reduction efficacy, it is crucial to use appropriate scaling factors in quantized representations of neural networks. That is, assuming an integer aligned uniform quantization scheme, the integer values are multiplied by a scalar step size to better capture the underlying data distribution. The work in~\cite{choukroun2019low} quantizes convolutional layers with one such step size per kernel and dense layers with a global step size. The step sizes are computed by a grid search, minimizing a mean squared error function. The authors showed that, given a calibration data set, stochastic gradient descent (SGD) can further refine the step sizes, improving $\textrm{top-}1$ accuracy by up to $23\%$.
The works in~\cite{Bhalgat_2020_CVPR_Workshops, jung2019learning} present direct step size learning in the scope of so-called \textit{quantization-aware training} on a per layer granularity. As a counterpart, in non-uniform quantization each quantization level can be learned by a corresponding scaling factor independently, i.e., they are not equidistantly distributed.  This technique finds application in ternary networks~\cite{Marban_2020_CVPR_Workshops}, i.e., having only two trainable quantization levels (``\textit{centroids}'') per network layer, or generally for low-precision networks to fine-tune centroids~\cite{SongHanMD15}.
\textit{Local Scaling Adaptation} is a technique that was adopted into the international standard ISO/IEC $15938-17$ on neural network compression (NNC)~\cite{kirchhoffer2021overview, haase2021encoder} to increase the model capacity and thus compensate for quantization errors. It introduces multiplicative factors at each output element of NN layers, which are then applied to the quantized parameter representation. 

In the context of distributed learning, several works compensate quantization errors by keeping track of the difference between the original and quantized gradients~\cite{wu2018error}. The work in~\cite{zheng2019communication} partitions gradients into blocks, where each gradient block (i.e., each tensor in their example) is compressed and transmitted in a 1-bit format with a corresponding scaling factor.
Another method is to reduce the communication frequency and not send gradients each iteration but weight updates as generalized gradients after clients have performed multiple iterations, finally averaging the updates on the server side (\textit{FedAvg})~\cite{fedavg}. The work in~\cite{mills2019communication} demonstrates that also an \textit{Adam}~\cite{ADAM}-optimized variant of FedAvg converges faster than traditional SGD, however, at the cost of additionally transmitting per-parameter learning rates and estimates of the 1st and the 2nd raw moment of the gradient.
Gradient sparsification methods only send gradient elements larger than a predefined threshold. One of the first approaches was presented by~\cite{strom2015scalable}, using a predefined threshold, whereas in~\cite{LinHM0D18} a fixed sparsity rate is used. In practice, however, it is non-trivial to choose suitable thresholds or rates, as they can vary considerably for different neural architectures and even different layer types, layer locations within the neural network, or in different training iterations~\cite{AjiH17}. 
The architectures are predefined in most of the works, which may not be the optimal choice for the particular FL scenario. Beyond improving communication settings, \textit{FedNAS}~\cite{FedNAS} proposes a paradigm for personalized FL by adapting client model architectures.

Our approach exploits the effects of parameter scaling in FL scenarios to speed up model convergence and reduce the amount of data to be transmitted. Scaling factors added at the granularity of convolutional filters allow overall sparser updates in the FedAvg paradigm in which we do not communicate additional Adam information, such as learning rates.
Together with a dynamic threshold for (structured) sparsification of weight updates, the scaling can faster adapt client networks to new data domains and implicitly personalize them by amplifying and suppressing their local filters. 

\section{Compression Pipeline For Differential Neural Network Updates}
\label{sec:compression}

In this work, one communication epoch $t$ is defined as follows: 1) clients are synchronized with the server, i.e., they download the latest server update and add it to their model state, 2) clients optimize weights based on their local data, 3) clients compute the weight difference $\Delta\mathcal{W}$ wrt. the previous state:
\begin{equation}\label{eq:w_update}
\Delta\mathcal{W}^{(t)}_i = \mathcal{W}^{(t)}_i - \mathcal{W}^{(t-1)}_i
\end{equation}
with $i \in I = [1, \ldots,\#\text{clients}]$,
4) compression of $\Delta\mathcal{W}^{(t)}_i$, 5) clients send the compressed $\Delta\hat{\mathcal{W}}^{(t)}_i$ to the server, 6) the server $\mathbb{S}$ applies federated averaging: $\Delta\mathcal{W}^{(t)}_{\mathbb{S}}~=~\frac{1}{\vert I\vert}\sum_{i \in I} \Delta\hat{\mathcal{W}}^{(t)}_i$, 7) the server sends the update $\Delta\mathcal{W}^{(t)}_{\mathbb{S}}$ to clients for synchronization, and so on. This basic communication protocol is in the spirit of federated averaging (FedAvg)~\cite{fedavg}. Unless otherwise specified, no error accumulation is used.

An integral aspect of our proposed method is the compression of weight updates, which is typically not part of FedAvg. In a first preprocessing step, sparsification techniques are applied which set unimportant weight update elements to zero, resulting in sparser and more compressible  tensors. \textit{Importance} in this context describes the impact a particular weight update element has on the computed output values of the model. We use the magnitude of the weight update elements as a heuristic for importance.

Sparsification can be carried out in an unstructured or structured manner. In unstructured sparsification, any weight update element with small magnitude is set to zero, independently of its position.
In contrast, structured sparsification sets an entire regular subset of parameters to zero, e.g.,  convolutional  filters, kernels, matrix rows or columns. In our proposed method we use both paradigms, structured and unstructured sparsification. For unstructured sparsification we calculate the threshold $\theta_{\textrm{u}}$ parameter-wise by Gaussian approximation, i.e.,

\begin{align}\label{eq:gaussapprox}
\notag \theta_{\textrm{u}} = \max(&\vert \textrm{mean}(\Delta W) - \delta \cdot \textrm{std}(\Delta W)\vert, \\
&\vert \textrm{mean}(\Delta W) + \delta \cdot \textrm{std}(\Delta W)\vert)\\
& \notag \textrm{s.t.} \quad \theta_{\textrm{u}} \geq \textrm{step\_size} / 2.
\end{align}

Here $\Delta W$ corresponds to a specific layer/parameter update within the neural network update $\Delta\mathcal{W}$, std($\cdot$) describes the standard deviation and $\delta$ is a hyperparameter which shifts the threshold and can be fine-tuned until a certain amount of sparsity or model performance degradation is exceeded. The step\_size is a global parameter used for quantization. More precisely, in our uniform quantization scheme, it is an integer range multiplied by a float value which is used to generate quantization levels to which the original weight update distribution is assigned to, i.e., $[-q, \ldots,-2, -1, 0, 1, \ldots, p] \cdot \text{step\_size}, \ q, p \in \mathbb{N}$.

For structured sparsification, we use convolutional filters~$F \in \mathbb{R}^{N\times K \times K}$ and output neurons $O \in \mathbb{R}^N$ as regular subsets of weight update elements when considering convolutional layers~$W_{\textrm{conv}} \in \mathbb{R}^{M\times N \times K \times K}$ and dense layers~$W_{\textrm{dense}} \in \mathbb{R}^{M\times N}$, respectively.
Here, $N$ refers to the number of input channels/elements, $M$ to the number of output channels/elements and $K \times K$ to the convolutional kernel size.
For simplicity, we use the notation of filter sparsification, which in the following shall also include sparsification of output neurons in dense layer types. As a threshold for structured sparsification $\theta_{\textrm{s}}$, we calculate the average of filter update means, parameter-wise:
\begin{equation}
\label{eq:filtermean}
\theta_{\textrm{s}} = \frac{\gamma}{M}\sum_{m=0}^{M-1} \vert \Delta \bar{F}\vert, \quad F \in \mathbb{R}^{N\times K \times K}.
\end{equation}
$\gamma$  is a hyperparameter which shifts the threshold and can be fine-tuned. Structured sparsity can be exploited very effectively in coding mechanisms, e.g., by skipping matrix rows that belong to corresponding sparse filter updates. 

Having introduced unstructured and structured sparsity, the model update $\Delta \mathcal{W}$ is uniformly quantized and encoded with the NNC standard. 
Since the universal \textit{DeepCABAC} entropy encoder of NNC can represent frequently occurring symbols with fewer bits, a high sparsity rate, i.e., a high probability that a weight update element $\Delta w_{nkk}=0$, reduces the model's entropy and thus results in smaller representations of~$\Delta \mathcal{W}$.

\section{Filter and Output Neuron Scaling for\\Federated Learning Scenarios}
\label{sec:method}

In our proposed method, we introduce additional trainable scaling factors $\mathcal{S}$ to compensate sparse network updates $\Delta \mathcal{W}$. 
The trade-off is that the more zero elements are transmitted during a weight update, the less learning progress is possible because the current state of the model cannot change significantly. At the same time, we aim to have many zero elements to make the data more compressible. With our proposed scaling factors, we intend to balance this trade-off, i.e., we compensate for the fact that many update elements are zero by adjusting neighboring weights with a multiplicative factor, even if the associated neighborhood update is entirely zero.

The scaling factors are implemented at the granularity of convolutional filters/output neurons which ensures a low memory and computational overhead of the extra parameters. 
For the implementation, the computational graph of the neural network must be adapted by equipping convolutional and dense layers with a multiplication function and a multi-dimensional parameter whose first dimension’s size corresponds to the number of output elements of the associated layer, while all following dimensions are of size~$1$ (``unsqueezed'') to allow for correct tensor multiplication. The overall number of dimensions is equal to the associated layer's number of dimensions, e.g., for a convolutional layer $W_{\textrm{conv}}~\in~\mathbb{R}^{M\times N \times K \times K}$ the scaling factors are \mbox{$S \in \mathbb{R}^{M\times 1 \times 1 \times 1}$}, which allows each single filter $F \in \mathbb{R}^{N\times K \times K}$ in $W_{\textrm{conv}}$ to be multiplied by a scalar $s$: 
\begin{equation}
\label{eq:scaling}
F^*_m = F_m\cdot s_m, \quad m \in [0, \ldots, M-1].
\end{equation}
In the following, $s$ refers to a single scaling factor, i.e., $s~\subseteq~S~\subseteq~\mathcal{S}$.
Apart from the increased model capacity, the additional parameters lead to different mechanisms taking effect in the context of incremental weight update compression of NNs. 
Filter scaling speeds up the convergence of the center model, although very sparse weight updates typically require many communication rounds to converge. As shown in the experiments section, the scaling factors can not only amplify certain convolutional filters but also attenuate or even suppress redundant filters, i.e., feature extractors, and increase the overall sparsity of model updates~$\Delta \mathcal{W}$. 
\begin{algorithm}[t]
\small
\DontPrintSemicolon
\SetKw{KwIni}{init:}
\caption{Unidirectional filter-scaled sparse federated learning (\textbf{FSFL})}\label{alg:one}
\textbf{input} \ initial model $\mathcal{W}^{(0)}$, dataset $\mathcal{D}$, $\#$clients, T, E\;
\textbf{output} \ federated learned server model $\hat{\mathcal{W}}^{(T)}_{\mathbb{S}}$\;
\textbf{init} \ all clients $\mathbb{C}_i , i \in I = [1, \ldots,\#\text{clients}]$ are
initialized with the server $\mathbb{S}$ parameters $\mathcal{W}_i \gets \mathcal{W}_{\mathbb{S}}$. Every client holds a different data split $\mathbb{C}_i \gets \mathcal{D}_i$, $\mathcal{D}_{\textup{val}_i}$. Scaling factor initialization is $\mathcal{S}_i \gets 1$, $\mathcal{S}_i \subseteq \mathcal{W}_i$. \\
\For{$t = 1,\ldots, T$}{
    \textit{Client side training:}\;
  \For{$i \in I$}{
  download$_{\mathbb{C}_i \gets \mathbb{S}}(\Delta\mathcal{W}^{(t)}_{\mathbb{S}})$ \;
  $\mathcal{W}^{(t)}_{i} = \mathcal{W}^{(t)}_{i} + \Delta\mathcal{W}^{(t)}_{\mathbb{S}}$\;
  $\mathcal{W}^{(t+1)}_i = $ train$(\mathcal{W}^{(t)}_{i}, \mathcal{D}_i ), \ \text{s.t.} \ \mathcal{S}^{(t+1)}_i = \mathcal{S}^{(t)}_i$\;
  $\Delta^s\mathcal{W}^{(t+1)}_i = $ sparsify$(\mathcal{W}^{(t+1)}_i - \mathcal{W}^{(t)}_i, \mathcal{D}_i)$\;
  $\hat{\mathcal{W}}^{(t+1)}_i = \mathcal{W}^{(t)}_i + \Delta^s\mathcal{W}^{(t+1)}_i$\;
  $\mathcal{S}^{(1)}_i \gets \hat{\mathcal{W}}^{(t+1)}_i$, \ perf $=$ eval$(\hat{\mathcal{W}}^{(t+1)}_i, \mathcal{D}_{\textup{val}_i})$\;
    \For{$e = 1,\ldots, E$}{
    $\mathcal{S}^{(e+1)}_i = $ train$(\mathcal{S}^{(e)}_i, \mathcal{D}_i)$\;
        \If{\textup{eval}$(\mathcal{S}^{(e+1)}_i, \hat{\mathcal{W}}^{(t+1)}_i, \mathcal{D}_{\textup{val}_i}) \geq$ \textup{perf}}{
        perf = eval$(\mathcal{S}^{(e+1)}_i, \hat{\mathcal{W}}^{(t+1)}_i, \mathcal{D}_{\textup{val}_i})$\;
        $\hat{\mathcal{W}}^{(t+1)}_i \gets \mathcal{S}^{(e+1)}_i$\;
        }
    }
    $\Delta\hat{\mathcal{W}}^{(t+1)}_i = \hat{\mathcal{W}}^{(t+1)}_i - \mathcal{W}^{(t)}_i$\;
    upload$_{\mathbb{C}_i \to\mathbb{S}}(\Delta\hat{\mathcal{W}}^{(t+1)}_i)$\;
  }
  \textit{Server side aggregation:}\;
  $\Delta\mathcal{W}^{(t+1)}_{\mathbb{S}} = \frac{1}{\vert I\vert}\sum_{i \in I} \Delta\hat{\mathcal{W}}^{(t+1)}_i$\;
  $\hat{\mathcal{W}}^{(t+1)}_{\mathbb{S}} =\mathcal{W}^{(t)}_{\mathbb{S}} + \Delta\mathcal{W}^{(t+1)}_{\mathbb{S}}$\;
}
\Return{$\hat{\mathcal{W}}^{(T)}_{\mathbb{S}}$}
\end{algorithm}

Algorithm~\ref{alg:one} gives an overview of our proposed \textit{filter-scaled sparse federated learning (FSFL)} scheme. First, clients $\mathbb{C}_i$ are synchronized with the server $\mathbb{S}$ by receiving its averaged update, then each client is trained on its local data split $\mathcal{D}_i$. Note that in this training step the scaling factors  $\mathcal{S}_i$ remain unchanged. After optimizing the clients, the differences wrt. the previous state are calculated and sparsified according to Equations (\ref{eq:gaussapprox}) and (\ref{eq:filtermean}). The sparsified updates $\Delta^s\mathcal{W}^{(t+1)}_i$ are then added to the previous model state $\mathcal{W}^{(t)}_i$ which will serve as a basis for scaling factor optimization. For $E$ sub-epochs, $\mathcal{S}_i$ are trained, whereas the rest of the network is frozen, including the running variances and means of the BatchNorm modules. After each sub-epoch, the scaled network is evaluated on the validation data set of the respective instance $\mathcal{D}_{\textup{val}_i}$. From all sub-epochs, the network variant with the best validation performance is selected and proceeded with. 

If the scaling factor training does not improve model performance compared to the sparsely updated model $\hat{\mathcal{W}}^{(t+1)}_i$, the scaling factor updates are discarded. Otherwise, the parameter differences of the model are recalculated (considering also the scaling factor parameters), quantized and encoded prior to transmitting them to the server. Note that we also investigated up- \textit{and} downstream compression, where the server update $\Delta\mathcal{W}^{(t+1)}_{\mathbb{S}}$ is sparsified and scaled as well. This is not shown in Algorithm~\ref{alg:one}.  

\subsection{Learning Rate Scheduling for Scaling Factors}
\label{ssec:warmrestart}

\begin{figure}[t]
\center
\includegraphics[width=1.0\linewidth]{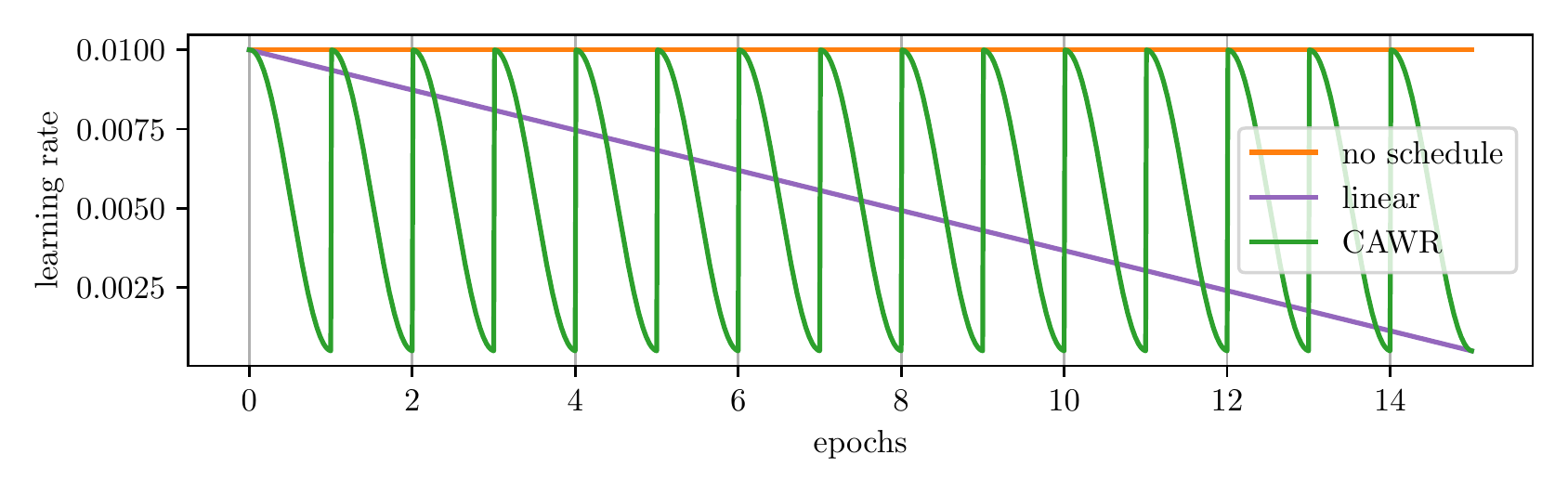}
\caption{Learning rate schedules used with $|T|=$15 epochs.}
\label{fig:cos}
\end{figure}

In our studies, we tested different optimizer-learning-rate-scheduler combinations, since the learning rate is crucial for appropriate convergence of the defined optimization problem. Even though the Adam optimizer~\cite{ADAM} supports individual adaptive learning rate estimates for different parameters, the resulting learning rates are sometimes unsuitable for gradient optimization of our present FL scenarios. If the initial rate is too small, the optimizer is likely to converge to local minima and does not improve performance while simultaneously increasing the bitstream sizes due to the additional scaling factors. On the other hand, larger rates lead to coarser steps in gradient descent and thus are suboptimal for loss optimization in the later epochs. 

To limit Adam’s peak learning rate, a linearly decreasing learning rate schedule can improve the results among others (e.g., the cosine annealing learning rate scheduler with warm restarts (CAWR) as proposed in~\cite{SGD_warm_restarts}). The schedulers are implemented such that after each batch of inferenced data, the scheduler performs one step according to the functions shown in  Figure~\ref{fig:cos}. The warm restarts of CAWR are introduced after each main training epoch $t$, prior to training the scaling factors $\mathcal{S}$. We also ran experiments using SGD for scaling parameter optimization.

\section{Experiments}
\label{sec:exp}

In the experiments, we evaluate our novel compression pipeline for FL scenarios by deploying three widely used neural network families (MobileNet, ResNet, VGG) and three datasets which are introduced in this section. We investigated different optimization methods, communication protocols, scaling factor parameterizations and their effects on weight updates as well as computational overhead, variation in number of clients and comparison with prior work.

\subsection{Experimental Setup}
\label{ssec:setup}

\begin{figure*}[!ht]
\centering
\includegraphics[width=0.495\textwidth]{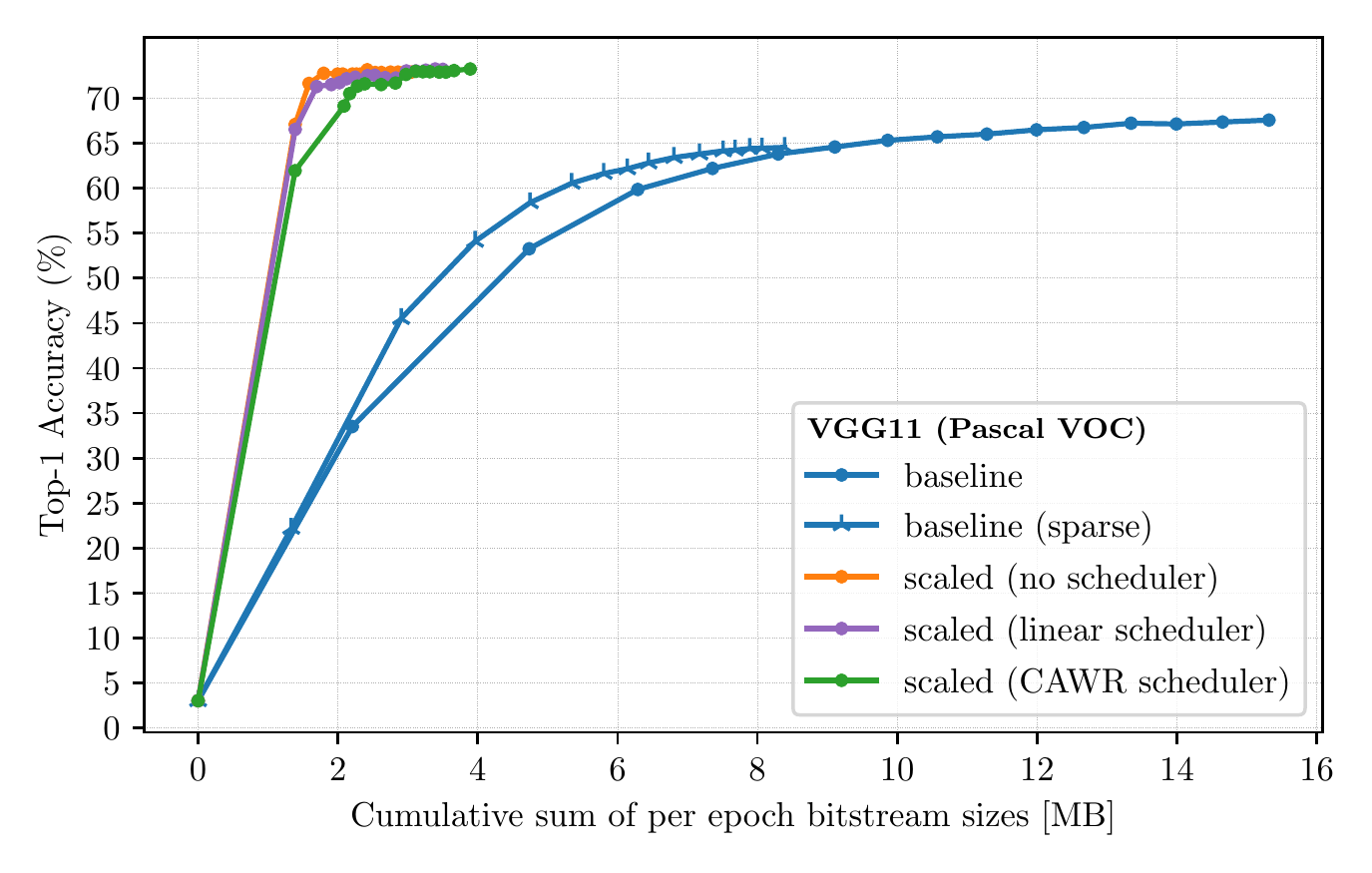}
\includegraphics[width=0.495\textwidth]{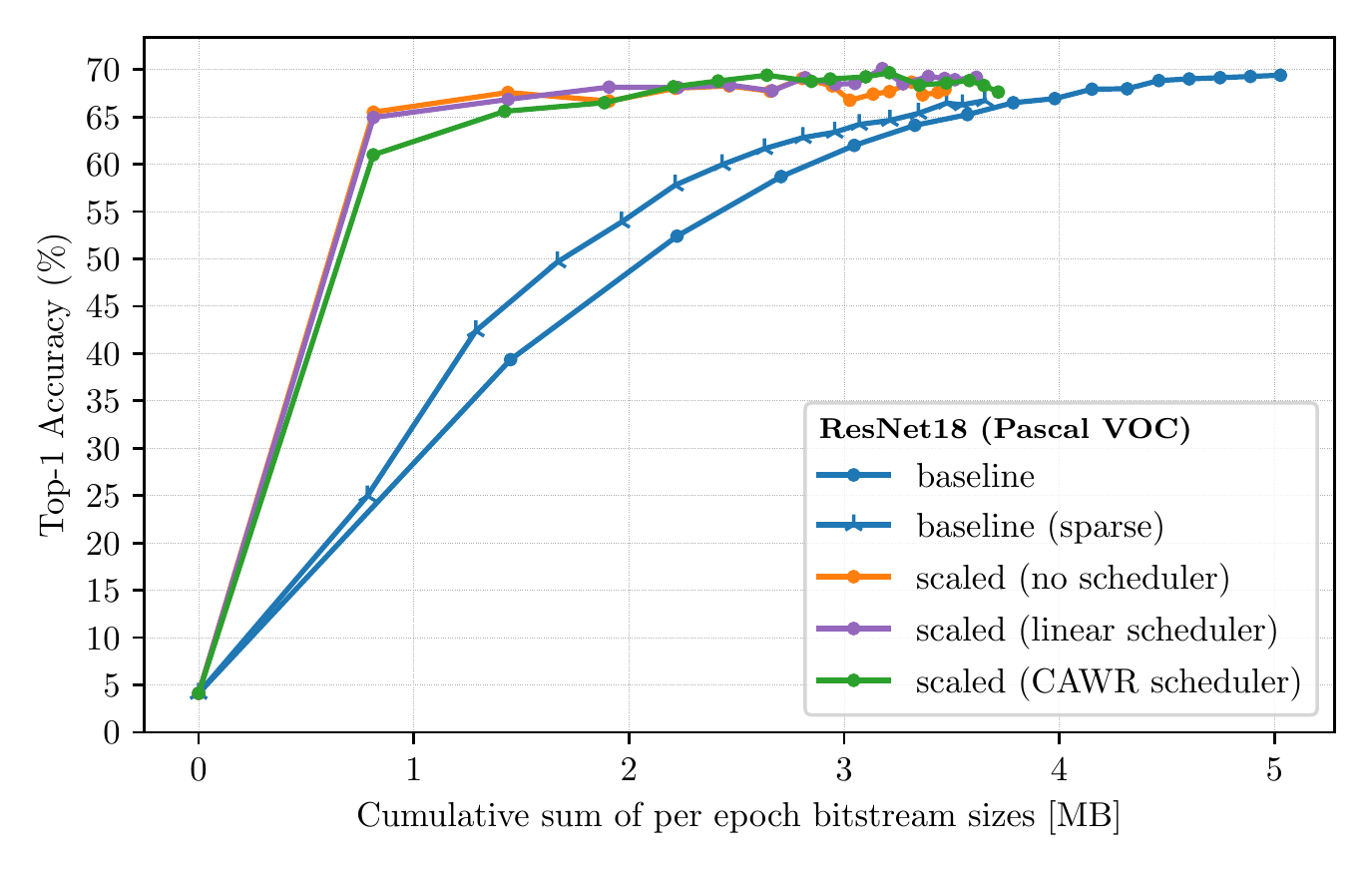}
\includegraphics[width=0.495\textwidth]{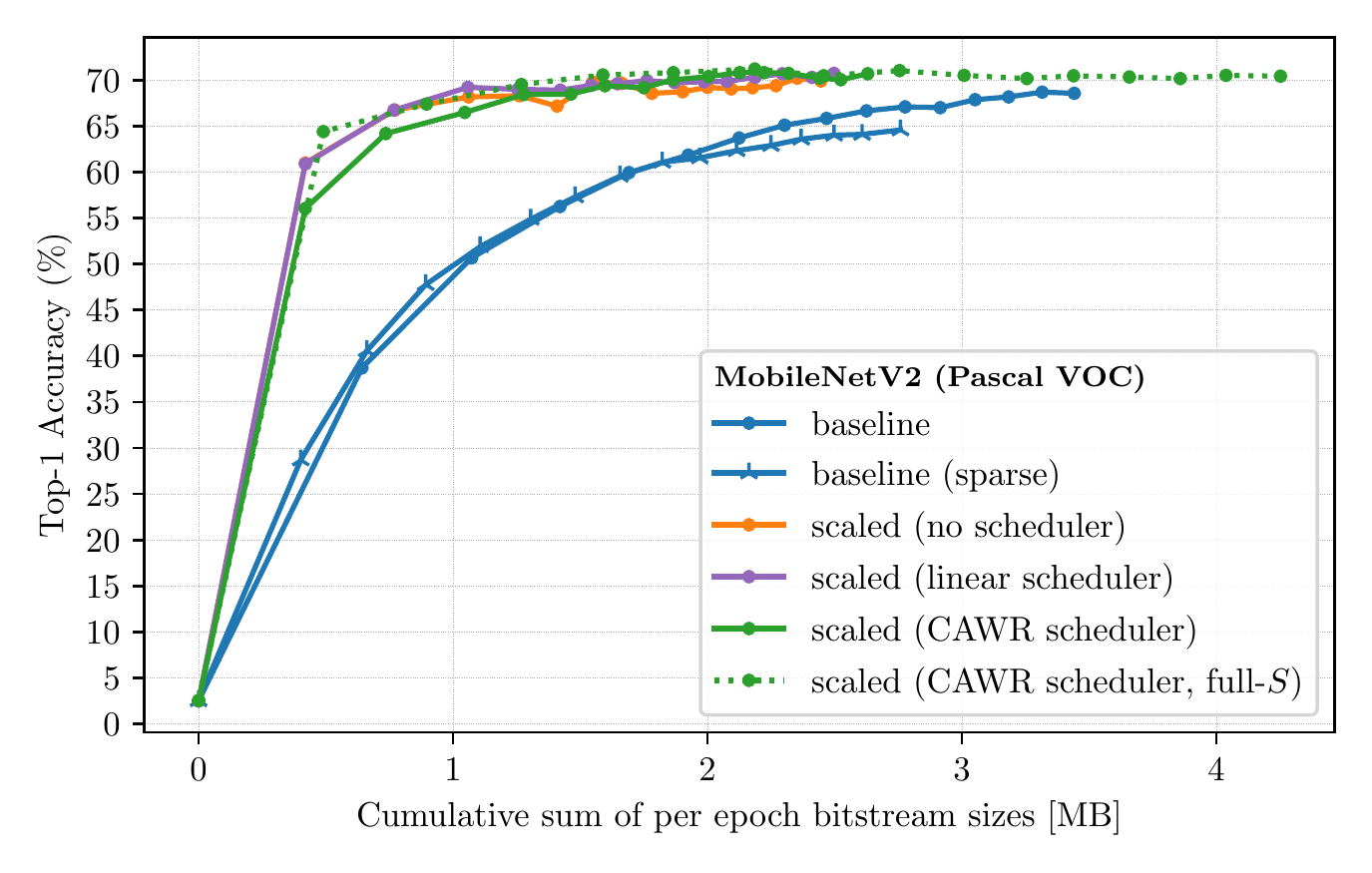}
\includegraphics[width=0.495\textwidth]{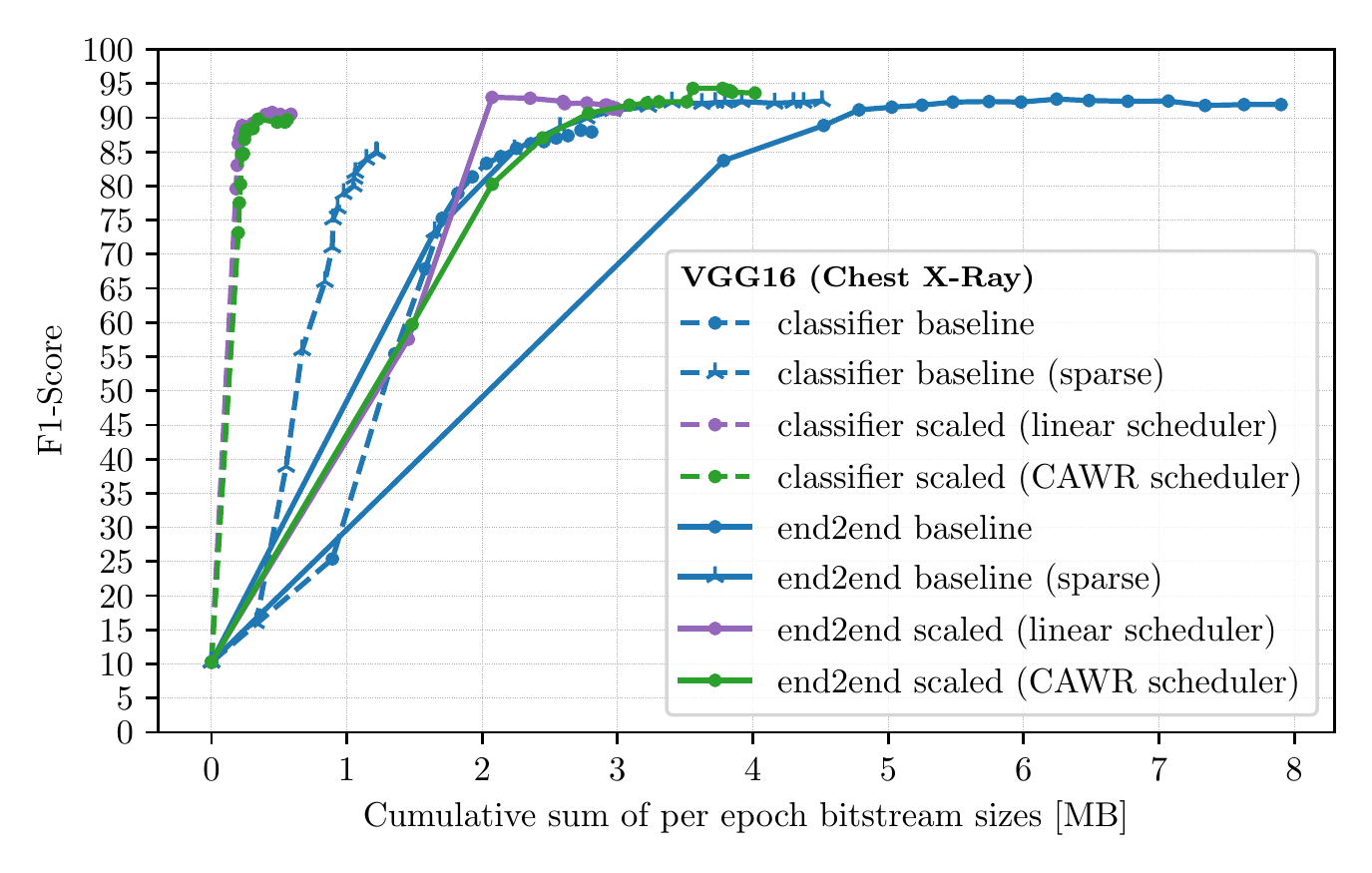}
\caption{Filter-scaled sparse federated learning (FSFL) of VGG11, ResNet18, MobileNetV2 and VGG16 (row-major order) solving the Pascal VOC and Chest X-Ray classification tasks, respectively. The MobileNetV2 chart includes experiments with scaling factors attached only to the output convolutions of each inverted residual block, compared to one experiment with full scaling factors (full-$S$). The VGG16 results include bidirectional compression of partial and end-to-end updates.}
\label{fig:mainres}
\end{figure*}

The used neural networks are pre-trained on ImageNet and downloaded from the torchvision model zoo\footnote{\label{torchvision}\url{https://pytorch.org/vision/stable/models.html}}. We adapted the classifiers to predict $20$, $10$ and $2$ instead of $1,000$ classes for the computer vision datasets \textit{Pascal VOC}~\cite{pascal-voc-2012}, \textit{CIFAR10}~\cite{cifar10} and  \textit{Chest X-Ray}~\cite{kermany_labeled_2018}, respectively.  Training and validation data were randomly split into non-overlapping client data sets $\mathcal{D}_i$. Note that for the sake of generalization, the optimization of scaling factors is evaluated locally using the validation data splits $\mathcal{D}_{\textrm{val}_i}$ and finally tested at the server's global model on the test data split.
The batch sizes used are 64 for VOC and CIFAR and 100 for Chest X-Ray. The VGG11 network to solve the CIFAR task, named VGG11$_{\textrm{CIFAR10}}$ in the following, was thinned to $[32, 64, 128, 128, 128, 128, 128, 128]$ convolutional filters and $128$ input neurons in the dense layers.

Unless otherwise specified, model weights $\mathcal{W}$ are transfer-learned in the FL scenarios as described in Sections~\ref{sec:compression}-\ref{sec:method} for $|T|=15$ epochs using Adam with an initial learning rate of $1e-5$. 
Since we deploy $\leq 16$ clients and assume reliable communication with no specific hardware limitations, our setups fall into the \textit{cross-silo} FL category.

For uniform quantization of weight updates $\Delta \mathcal{W}$, we use a step\_size of $4.88e-4$ and $2.44e-4$ for uni- and bidirectional FL settings, respectively. Scaling parameter, bias and BatchNorm parameter updates are quantized with a step\_size of $2.38e-6$. For further details on software implementation and data splits, we refer to the Appendix \ref{appendix:details}.

\subsection{Results for Different Optimization Schedules}
\label{ssec:rdopt}


Figure~\ref{fig:mainres} depicts the federated learning process in terms of server model performance and overall transmitted data between clients and server. Specifically, each data point in the charts represents one round of communication and indicates 1) how much data in bytes has overall (accumulative) been transmitted since the last round and 2) which performance (accuracy, $\textrm{F1 score}$) has been achieved by the aggregated server model. 
Thus, our goal is to shift the curves as far as possible to the upper left corner of the charts to achieve fast and communication-efficient FL.

We compare different configurations: The baseline configuration neither includes filter scaling nor sparsification, the sparse baseline includes sparsification only and all other curves represent our FSFL method of scaled and sparse differential filters with various optimizers (Adam, SGD) and learning rate schedulers (no schedule, linear, CAWR). 

It is evident that training converges significantly faster when applying filter scaling. Also, the cumulative sum of transmitted data is reduced, which leads to significant savings in data traffic, especially for VGG11. The Adam optimization of scaling factors outperforms SGD optimization in all cases, so we refer to the Appendix \ref{appendix:sgd} for SGD results. For Adam optimization, linear and CAWR schedules improve the training process in different regimes: in later epochs, the use of CAWR often generates models with higher performance, however, in the earlier epochs a linear schedule achieves better $\textrm{top-}1$ accuracy/F1 scores.

In literature, it has been shown that it is meaningful to compress weight updates in a bidirectional fashion, i.e., center-to-clients and clients-to-center communication. We tested this scenario with the Chest X-Ray dataset as it is motivated by a possible real world scenario, where a number of hospitals jointly train the detection of pathological evidence in x-ray data and a central server regularly updates the local detectors in the hospitals. As can be seen in Figure~\ref{fig:mainres} (bottom right), also in the bidirectional compression scenario filter scaling can contribute to faster convergence, bitrate savings and improved model performance. And finally, we also compare a full model update (``end2end'') with a partial update which only updates the classifier part of the VGG16 network consisting of a BatchNorm module and two dense layers. Here, only 258 scaling factors were applied, still the improvements are substantial.

The achieved gains seem to be counter-intuitive at first glance, as additional scaling parameters (with a fine quantization level, i.e., less lossy) have been added to the encoded bit stream and still the total bit rate has decreased. Thus, this characteristic is closer examined in the following section.

\begin{figure}[!th]
\centering
\includegraphics[width=0.995\linewidth]{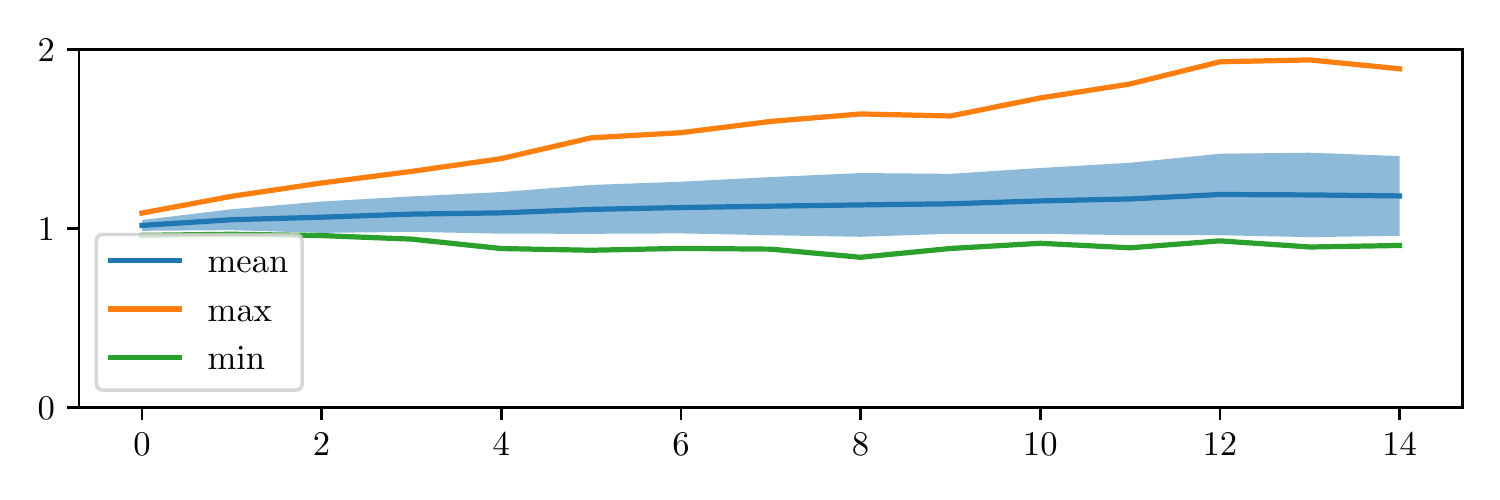}
\includegraphics[width=0.995\linewidth]{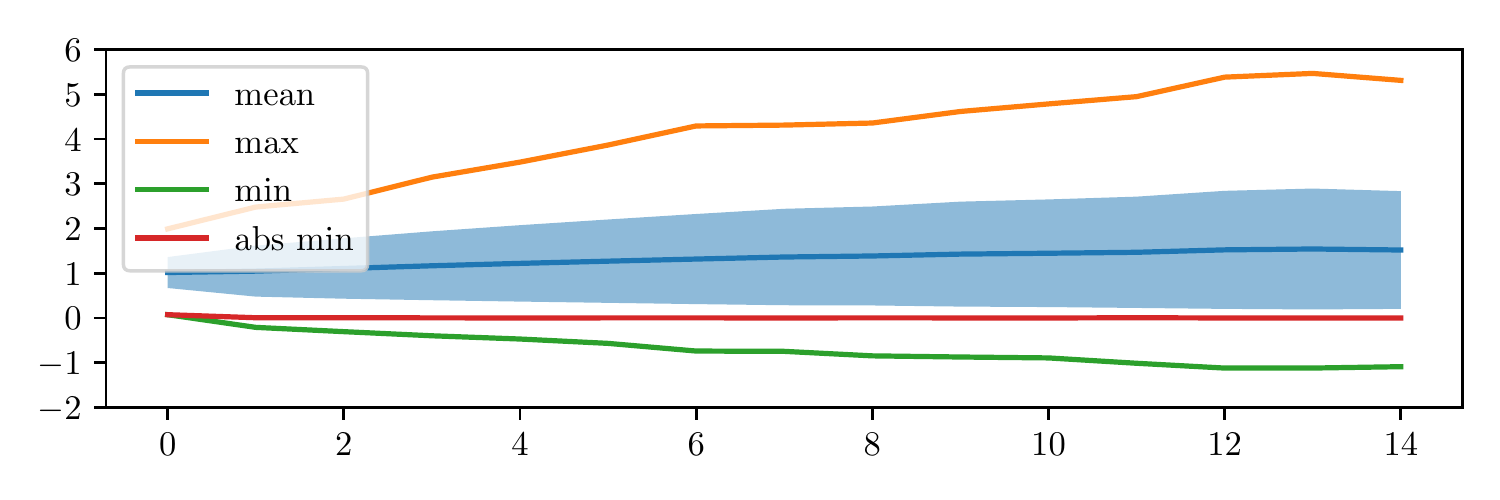}
\includegraphics[width=0.995\linewidth]{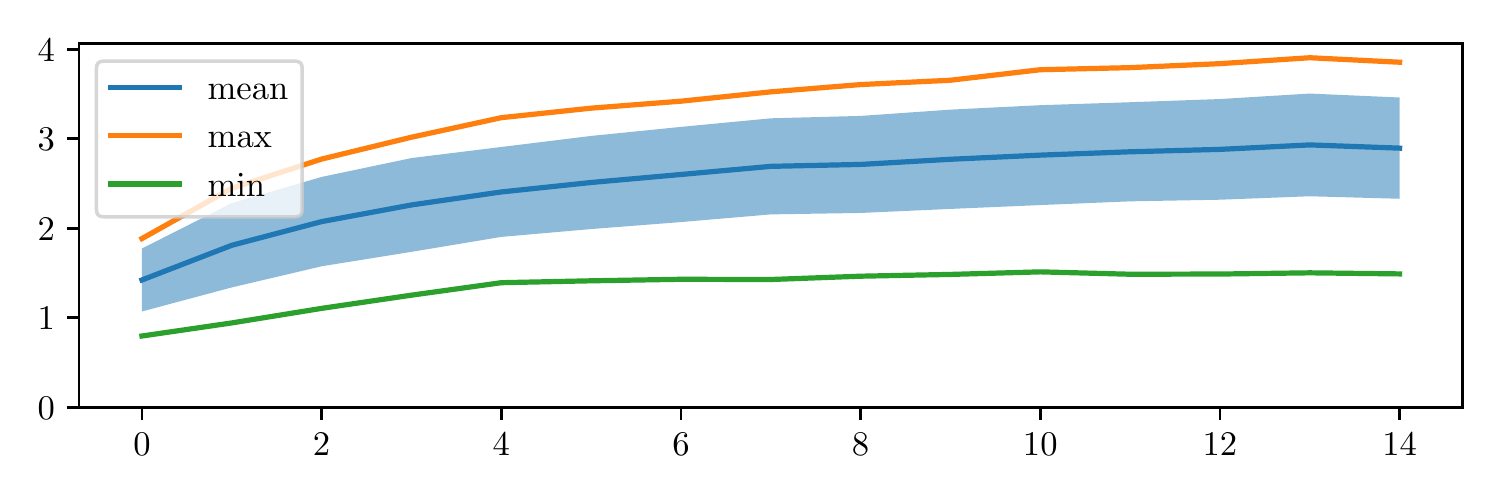}
\caption{Scaling factor statistics for MobileNetV2 during 15 epochs of training of three layers at different network positions: first inverted residual block, 17th inverted residual block, and the dense output layer (from top to bottom).}
\label{fig:scale_stats}
\end{figure}

\subsection{Effect of Filter-Tuning on Weight Updates}
\label{ssec:effect}

The scaling factors can amplify or suppress the impact of specific filters on the overall loss function, e.g., if a scaling factor $s$ is close to zero, the entire convolutional filter only marginally contributes to the net output as the generated feature map will be quite sparse or of very low magnitude. 
This also has a large impact on the weight updates $\Delta \mathcal{W}$ and can be particularly advantageous for FL in computer vision applications to eliminate redundant or unused feature extractors learned in divergent image domains. 

Having a closer look at Figure~\ref{fig:scale_stats}, we can observe different behaviors of scaling factors $\mathcal{S}$  dependent on their location within the neural network. On one hand, scaling factors in more shallow layers (close to the input) tend to converge to values close to $1$, i.e., they do not change much and thus only marginally influence the computational graph. In deeper located layers, the scaling factors on the other hand converge to values of $s\to 6$ while simultaneously suppressing other filters with values close to zero. Interestingly, in the dense output layer (bottom subplot of Figure~\ref{fig:scale_stats}) all output neurons are somehow amplified due to scaling.

\begin{figure}[t!]
\center
\includegraphics[width=0.995\linewidth]{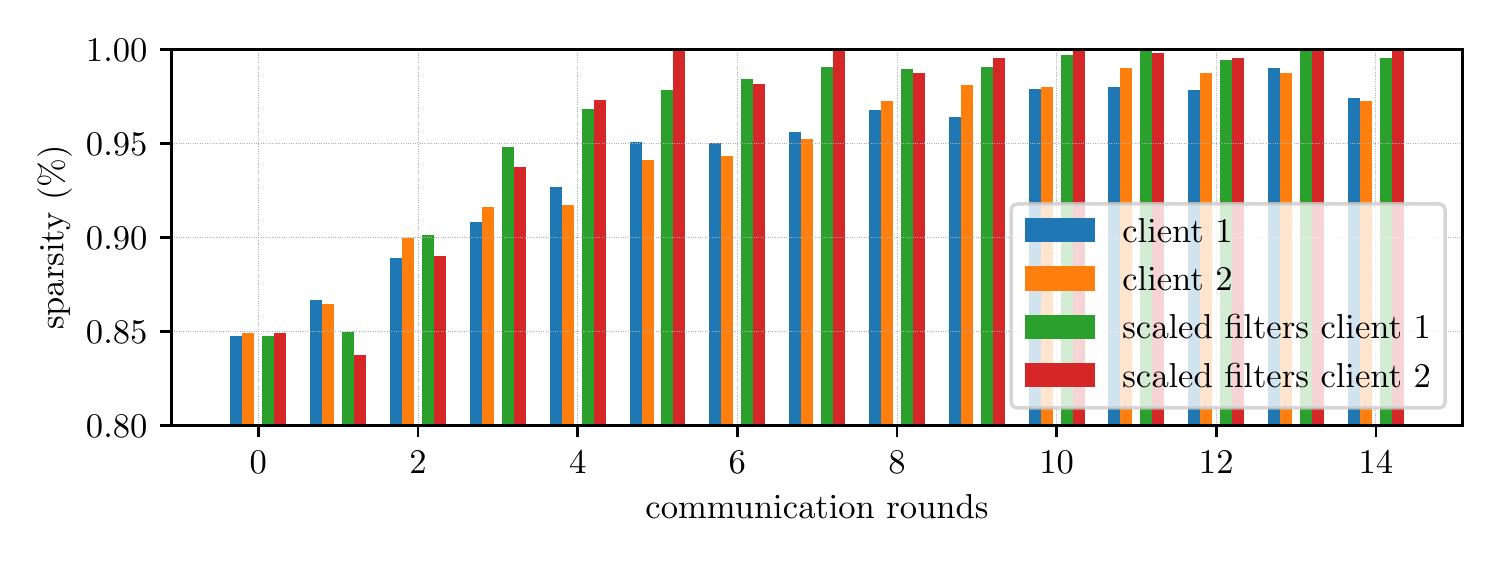}
\caption{Sparsity of two scaled \& unscaled MobileNetV2 clients.}
\label{fig:sparsity}
\end{figure}

A positive side-effect of these mechanisms is that also the filter weight updates $\Delta\mathcal{W}$ result in sparser tensors when using trainable scaling factors $\mathcal{S}$. It is not surprising that filters $F$ with corresponding $s\to 0$ generate sparse output features. Consequently, the backpropagation algorithm computes smaller gradients for the associated weights. Another effect might be that optimized $s$ make further training of filter weights redundant as the loss wrt. the filter $F$ has already converged due to ``macro training'' through~$s$. 

Figure~\ref{fig:sparsity} illustrates the sparsity of two clients per epoch when filter scaling and sparsification are applied vs. sparsification only. In the first few epochs, $\Delta\mathcal{W}$ is equal to or sparser than  $\Delta\mathcal{W}_{\text{scaled}}$, however in most of the epochs filter scaling increases sparsity, even up to $100\%$ (meaning that \textit{exclusively} scaling factors $\mathcal{S}$ are sent by the client, i.e., macro training only).
In summary, even though additional scaling factors are added to the transmitted bit streams, their impact increases the sparsity of filter updates, resulting in a final overall data reduction.

\subsection{Computational and Memory Overhead}
\label{ssec:overhead}

\begin{table}[t!]
\centering
\caption{Number of additional parameters and training time.}
\label{table:times}	
\resizebox{0.7125\linewidth}{!}{%
\begin{tabular}{l c c c} 
 \hline 
 model & $\#$params$_{\text{orig}}$ & $\#$params$_{\text{add}}$ & t$_{\text{add}}$ \\ [0.5ex] 
 \hline\hline
 MobileNetV2 & 2.3M & 2,836 & 1.17$\times$ \\ 
 with full-$\mathcal{S}$      & 2.3M & 17,076 & 1.31$\times$ \\ 
 ResNet18 & 11.2M & 4,820 & 1.62$\times$ \\
 VGG11 & 128.8M & 10,964 & 1.65$\times$ \\
  \hline
 VGG11$_{\textrm{CIFAR10}}$ & 0.8M & 1,002 & 1.68$\times$ \\
  \hline
 VGG16 & 16.8M & 4,482 & 1.60$\times$ \\
 VGG16$_{\textrm{partial}}$ & 2.1M & 258 & 1.40$\times$ \\
 \hline
\end{tabular}
}
\end{table}

In addition to the experimental results above, Table~\ref{table:times} gives an insight into the number of scaling parameters $\mathcal{S}$ and additional training time required. It shows that $\mathcal{S}$ only accounts for 0.009$\%$ to 0.748$\%$ of the total network parameters, i.e., the extra storage and size of the compressed update is small. The additional processing time of scaling parameter training does not directly scale with the total number of  $\mathcal{S}$. It also depends on the number of layers equipped with $\mathcal{S}$ and the overall size of the network. This is due to the fact that the optimization algorithm considers the whole computational graph of the network for gradient propagation, even if only $\mathcal{S}$ is updated. 
However, performing one training iteration to update $\mathcal{W}$ compared to performing two iterations, one for $\mathcal{W}$ and one for $\mathcal{S}$, requires on average only $1.17\times$ to $1.68\times$ the original computation time, which was also reflected in the run times of our experiments.
This extra effort can be interpreted as an upper-bound and there are options to minimize the effort, e.g., by 

1) applying scaling parameter training less frequently; 

2) equipping less layers with $\mathcal{S}$, e.g., as tested with MobileNetV2, where we only equipped the final convolutional layer of each inverted residual block instead of all convolutional layers therein (cf. ``full-$\mathcal{S}$'' in Figure~\ref{fig:mainres} and Table~\ref{table:times}); 

3) focusing $\mathcal{S}$ to be applied in deeper layers, as shown in partial updates of the VGG16 network, which converged as the end-to-end training counterpart (see also Section~\ref{ssec:effect}, where we showed that scaling factors do not change much in shallower layers); 

4) considering smaller training splits to train $\mathcal{S}$, i.e., not the full training split which is used to train $\mathcal{W}$.

\subsection{Increasing the Number of Clients}
\label{ssec:numclients}

\begin{figure}[!t]
\center
\includegraphics[width=0.995\linewidth]{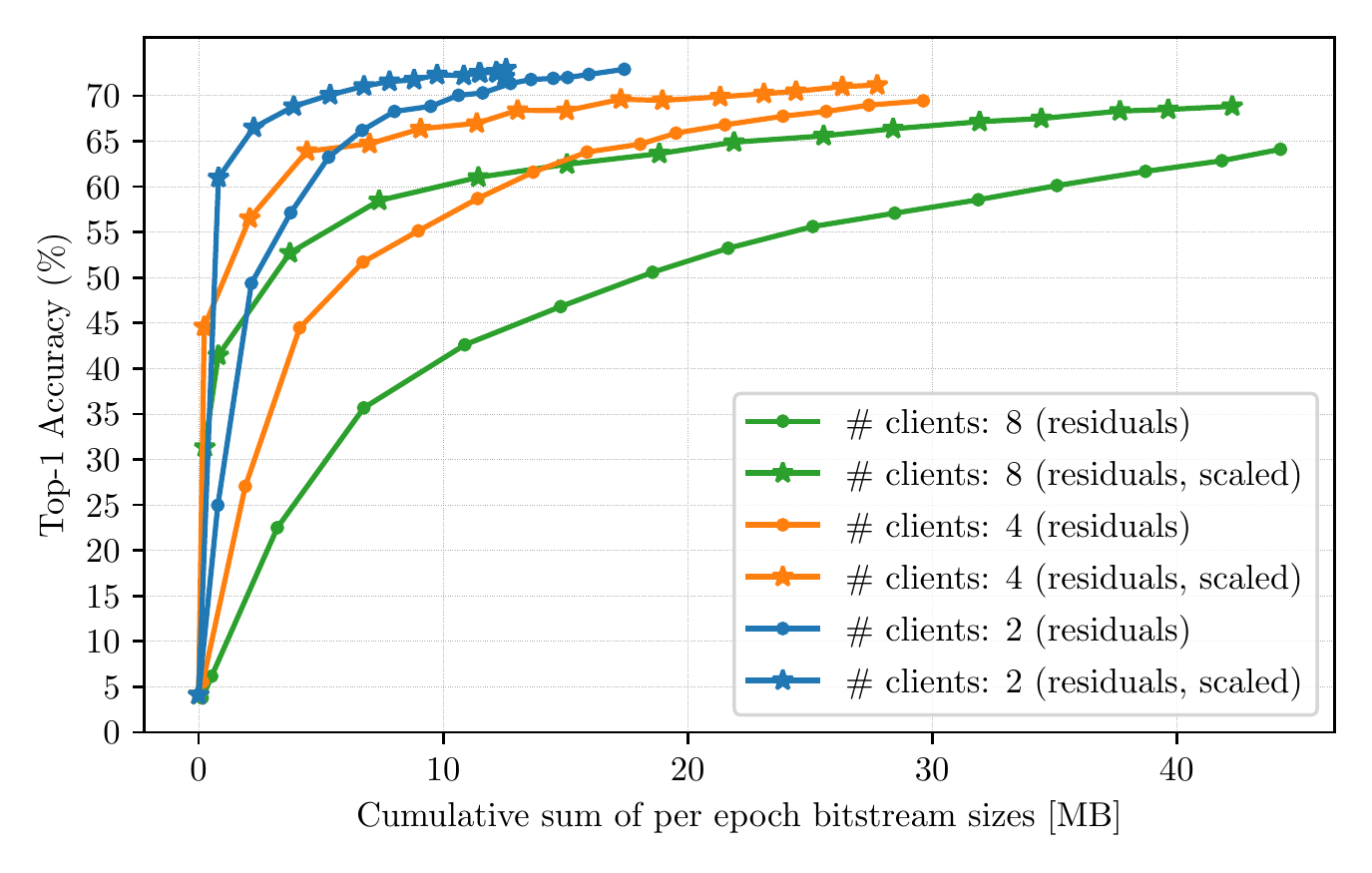}
\caption{ResNet18 with error accumulation (residuals) and variation in number of clients.}
\label{fig:multclients}
\end{figure}
The purpose of the following analysis is to investigate the potential scalability of our approach, i.e., whether filter scaling can still achieve fast and communication-efficient FL as the number of clients increases. In the Pascal VOC scenario, we increased the number of clients from $2$ to $4$~and~$8$. As this FL scenario with $\#\text{clients} \geq 4$ barely achieved representative accuracies \textit{without} our FSFL method, we implemented error accumulation as used in~\cite{sattler2019robust} after sparsification: 
\begin{equation}\label{eq:residual}
\Delta\mathcal{W}^{(t+1)}_i =  \mathcal{R}^{(t)}_i + \mathcal{W}^{(t+1)}_i - \mathcal{W}^{(t)}_i
\end{equation}
with $\mathcal{R}^{(t+1)}_i = \Delta\hat{\mathcal{W}}^{(t+1)}_i - \Delta\mathcal{W}^{(t+1)}_i$ and Equation~\ref{eq:residual} being inserted in Algorithm~\ref{alg:one}, line 10.
\begin{table*}[th!]
    \centering
    \caption{Comparing different approaches from literature and combinations with parts of our proposed compression pipeline with our filter-scaled and sparse FL method (FSFL). Experiments were conducted using a thinned VGG11 solving CIFAR10 with 2, 4, 8 and 16 clients for $t =1, ..., T=90$ epochs. A constant sparsity rate of $96\%$ was used for sparse ternary compression (STC) and our methods.}
    \label{tab:vs_stc}
    \resizebox{\linewidth}{!}{
    \begin{tabular}{l|cccc|cccc|cccc|cccc}
    \toprule
     & \multicolumn{4}{c|}{$|I|=2$ Clients} & \multicolumn{4}{c|}{$|I|=4$ Clients} &  \multicolumn{4}{c|}{$|I|=8$ Clients} &  \multicolumn{4}{c}{$|I|=16$ Clients} \\
    \midrule
    & \multicolumn{2}{c}{Acc. $=70.0$} &  \multicolumn{2}{c|}{Acc. $=\textbf{76.7}$}  & \multicolumn{2}{c}{Acc. $= 64.3$} &  \multicolumn{2}{c|}{Acc. $=\textbf{71.6}$}  & \multicolumn{2}{c}{Acc. $= 57.9$} & \multicolumn{2}{c|}{Acc. $=\textbf{67.2}$} & \multicolumn{2}{c}{Acc. $=48.2$} &  \multicolumn{2}{c}{Acc. $=\textbf{61.4}$} \\
    & $\sum$ data  & $t$ & $\sum$ data  & $t$ & $\sum$ data  & $t$ & $\sum$ data  & $t$ & $\sum$ data  & $t$ & $\sum$ data  & $t$ & $\sum$ data  & $t$ & $\sum$ data  & $t$ \\
    \midrule
    FedAvg~\cite{fedavg} & 387.13 MB & 58 & $\varnothing$ & $\varnothing$ & 747.09 MB & 56 & $\varnothing$ & $\varnothing$ & 1,467.02 MB & 55 & $\varnothing$ & $\varnothing$ & 2,988.37 MB & 56 & $\varnothing$ & $\varnothing$ \\
    FedAvg~\cite{fedavg}$^{\dagger}$ & 10.54 MB & 58 & $\varnothing$ & $\varnothing$ & 17.82 MB & 55 & $\varnothing$ & $\varnothing$ & 31.86 MB & 55 & $\varnothing$ & $\varnothing$ & 55.48 MB & 56 & $\varnothing$ & $\varnothing$ \\
    STC~\cite{sattler2019robust}$^{\dagger}$ & 4.33 MB & 73 & $\varnothing$ & $\varnothing$ & 8.65 MB & 74 & $\varnothing$ & $\varnothing$ & 16.56 MB & 72 & $\varnothing$ & $\varnothing$ & 34.11 MB & 76 & $\varnothing$ & $\varnothing$ \\
    Eqs. (\ref{eq:gaussapprox}) + (\ref{eq:filtermean}) & 3.71 MB & 90 & $\varnothing$ & $\varnothing$ & 7.23 MB & 90 & $\varnothing$ & $\varnothing$ & 14.24 MB & 90 & $\varnothing$ & $\varnothing$ & 27.46 MB & 90 & $\varnothing$ & $\varnothing$ \\
    STC~\cite{sattler2019robust}$^{\ddagger}$ & 1.97 MB & 31 & 5.53 MB & 86 & 4.34 MB & 34 & 10.59 MB & 83 & 7.81 MB & 31 & 21.79 MB & 86 & 10.63 MB & 22 & 42.81 MB & 86 \\
    \textbf{FSFL} & \textbf{1.68 MB} & 36 & \textbf{3.92 MB} & 90 & \textbf{3.61 MB} & 39 & \textbf{8.09 MB} & 90 & \textbf{6.16 MB} & 34 & \textbf{15.94 MB} & 90 & \textbf{7.93 MB} & 23 & \textbf{31.34 MB} & 90 \\
    \bottomrule
    \end{tabular}
    }
    \footnotesize{$^{\dagger}$ Literature method with DeepCABAC encoding. $^{\ddagger}$ Literature method with DeepCABAC encoding and our proposed filter scaling, cf. Equation (\ref{eq:scaling}).}\\
    \footnotesize{$\varnothing$ configuration has not achieved the target accuracy within 90 communication epochs}
\end{table*}

The described error accumulation stores the difference of the compressed update and the original full-precision update locally (``residual''). According to this scheme, also small update elements in terms of magnitude can sum up until they exceed a certain threshold. Compared to the experimental results without residuals (cf. ResNet18 in Figure~\ref{fig:mainres} vs. Figure~\ref{fig:multclients} with $\#$clients: 2), residuals produce updates with a higher bitrate, as more information is accessible and will be send. The performance of the center model is slightly increased, however if the number of clients increases, convergence speed decreases as a consequence of the higher degree of distribution in the system and the rising non-IID-ness in data due to random partitioning of client  data (see Appendix~\ref{appendix:datadistrib}). 
As our primary goal is faster convergence and bitrate reduction, we don't address robustness on non-IID data in the scope of this work. 

The results in Figure~\ref{fig:multclients} show that FSFL (``scaled'') again outperforms all unscaled training processes, which becomes even more evident as the number of clients increases. The relative improvement in $\textrm{top-}1$ accuracy is highest for the scenario with $8$ clients equally involved in federated learning, which may be a tentative indicator that the proposed method is scalable.

\subsection{Comparative Results}

After investigating several parameter variations of our FSFL method, this subsection provides final results in comparison to current state-of-the-art methods. Here, Federated Averaging (FedAvg~\cite{fedavg}) is a widely used baseline algorithm for communication-efficient FL. Accordingly, there are numerous works in the literature that propose improvements for FedAvg, of which Sparse Ternary Compression (STC~\cite{sattler2019robust}) is a regularly cited improvement~\cite{survey_fl_systemlevel,survey_fl_2}. STC converges faster when compared to other averaging algorithms like FedAvg while less bits are communicated overall. Because of these advantages, we also compare our method with STC. The STC protocol compresses communication via sparsification, ternarization, error accumulation - according to Equation (\ref{eq:residual}) - and Golomb encoding. For better comparability, and since DeepCABAC also makes use of Golomb encoding for binarization, we encoded weight updates with DeepCABAC in our STC implementation.

Table~\ref{tab:vs_stc} shows the respective experimental results. We deployed a thinned VGG11, as used in~\cite{sattler2019robust}, which was federally learned on $80\%$ of the CIFAR10 training data ($20\%$ were used as validation data). For the FL scenario, we used $2$, $4$, $8$ and $16$ clients which were trained for $T=90$ epochs with a constant sparsity rate of $96\%$. 

First, we executed FedAvg, where the uncompressed client model updates are sent after each epoch, then averaged on the server side and broadcast to the clients (here without server-to-clients compression, which holds for all results in Table~\ref{tab:vs_stc}). Second, we applied our proposed uniform quantization and DeepCABAC encoding to the \mbox{FedAvg} pipeline, which reduced the amount of communicated bits by a factor of $\approx54$. Third, STC is applied to the weight updates. 
Different from the experimental setup in~\cite{sattler2019robust}, we did not use an ``equivalent'' delay period of $n$ iterations for FedAvg, consequently STC introduces additional sparsity of weight updates, leading to convergence at later epochs $t$. 
In the fourth row of Table~\ref{tab:vs_stc}, we provide results for applying our proposed sparsification to the weight updates which further decreases the amount of communicated bits, however, at the cost of additional training epochs to converge. This can be explained due to the structured sparsity introduced in the weight updates which is not fine-tuned here but fixed to $96\%$ for better comparison. In a next step, we applied our filter scaling method to STC and to our proposed sparsification scheme (FSFL). 

With filter scaling enabled, both methods converge significantly faster, i.e., they require less communication rounds to achieve the target accuracies, while the amount of transmitted data is reduced by up to $\approx 377\times$ compared to FedAvg, and achieve higher $\textrm{top-}1$ accuracies overall compared to unscaled configurations.
In accordance with the previous section, the benefits of filter scaling increase as the population of clients grows. However, the accuracy achievable within 90 epochs, and thus the overall convergence speed, decreases with a larger number of participating clients (as also described in section~\ref{ssec:numclients}).

\section{Conclusion}
\label{sec:conc}

In this paper we presented a fast converging compression pipeline for FL scenarios in computer vision applications. The pipeline applies structured and unstructured sparsification, and equips weight layers with additional trainable scaling factors at the granularity of convolutional filters.
We showed that the scaling factors can amplify or suppress the impact of specific filters on the overall loss function and by doing so can compensate very sparse updates while improving convergence speed.
As a result, overall data and bit stream sizes are reduced. The proposed method was tested and verified in its data reduction capability with different FL settings, including a learning scheduler variation, single (server-to-clients) and double (also clients-to-server) communication, end-to-end and partial updates, as well as FL systems with different number of clients. 
Compared to previous work, the proposed scaling method converges faster, achieves higher accuracy and reduces the amount of total transmitted data by up to $\approx 377\times$.

\balance

{\small
\bibliographystyle{ieee_fullname}
\bibliography{refs}
}

\thispagestyle{empty}
\onecolumn
\appendix
\counterwithin{figure}{section}
\counterwithin{table}{section}
\nobalance

\section{Further Information on Software Implementation and Datasets}
\label{appendix:details}

In order to train scaling factors separately, an additional optimizer is instantiated, which only updates the scaling factors $\mathcal{S}$. All $s \in \mathcal{S}$ are initialized with the value~$1$ and then optimized with Adam or SGD with a momentum of~$0.9$.
From a software perspective, we implemented a wrapper function, which detects all convolutional and dense layers within the respective neural network and replaces them with a scaled version of the respective module class. Note that this procedure changes the computational graph, however scaling factors have no effect on the original graph output if set to~$1$.
All experiments were conducted on a homogeneous GPU cluster employing NVIDIA Ampere A100 GPUs (40 GB RAM). We use PyTorch 1.8.1 and torchvision 0.9.1 as deep learning framework and CUDA 11.1 for NN GPU acceleration.

The \textit{Pascal Visual Object Classes Challenge 2012}~\cite{pascal-voc-2012} provides $11,540$ images categorized into 20 classes. The \textit{Chest X-Ray} dataset~\cite{kermany_labeled_2018} consists of $5,856$ images which are categorized into two classes: ``pneumonia'' and ``normal''. \textit{CIFAR10}~\cite{cifar10} consists of $60,000$ images with a resolution of $32\times 32$ pixels, containing 10 classes.
The data sets are split into approximately $60:20:20 \%$, $75:15:10 \%$ and $70:15:15 \%$ for training:validation:testing purposes for Pascal VOC, Chest X-Ray and CIFAR10, respectively.
We applied normalization and random horizontal flipping to all samples. Additionally, VOC samples are center cropped to $224\times 224$ pixels and Chest X-Ray samples to $150\times 150$ pixels. 

\section{SGD Results}
\label{appendix:sgd}

\begin{figure}[h]
\centering
\includegraphics[width=0.495\textwidth]{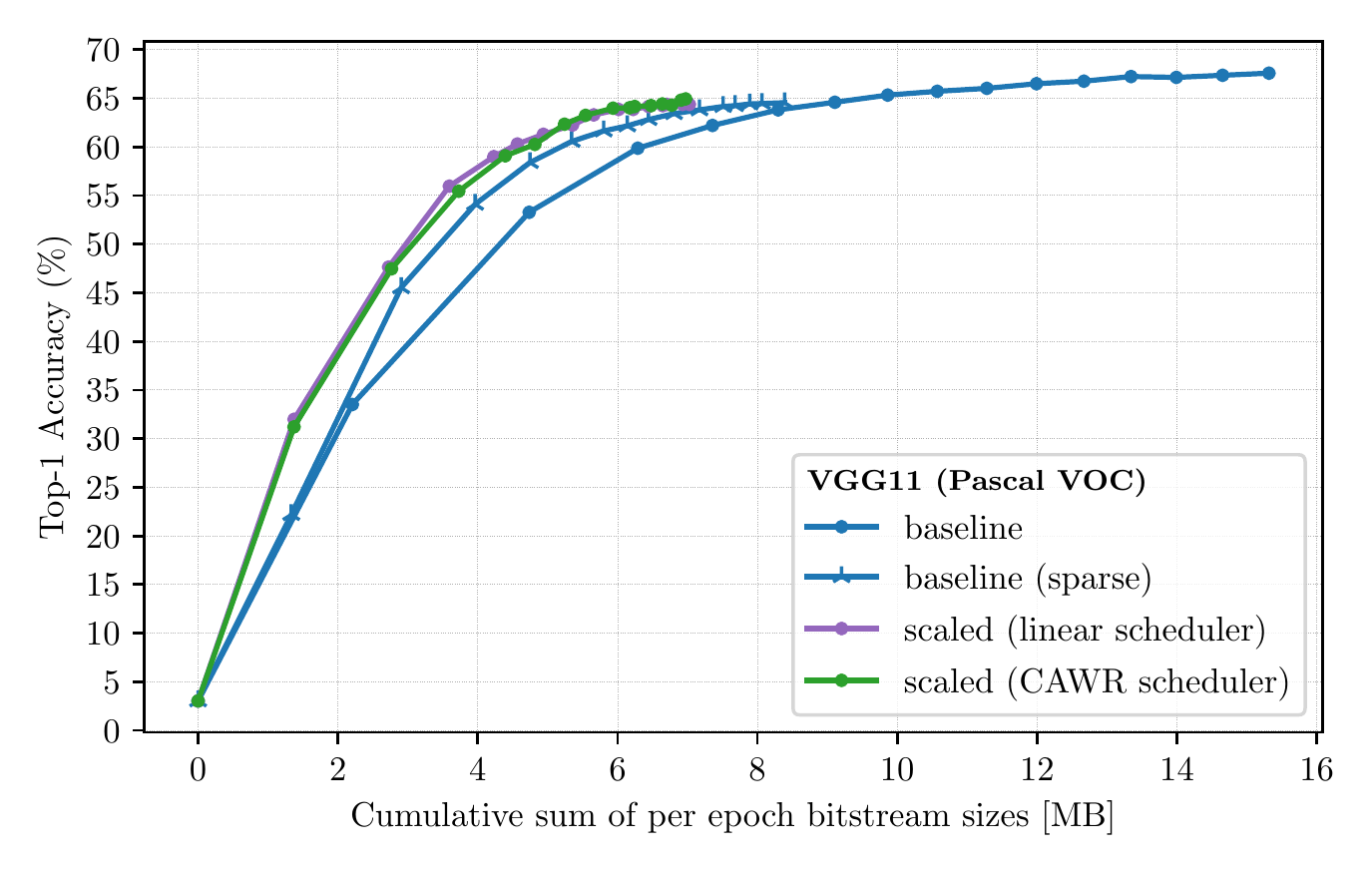}
\includegraphics[width=0.495\textwidth]{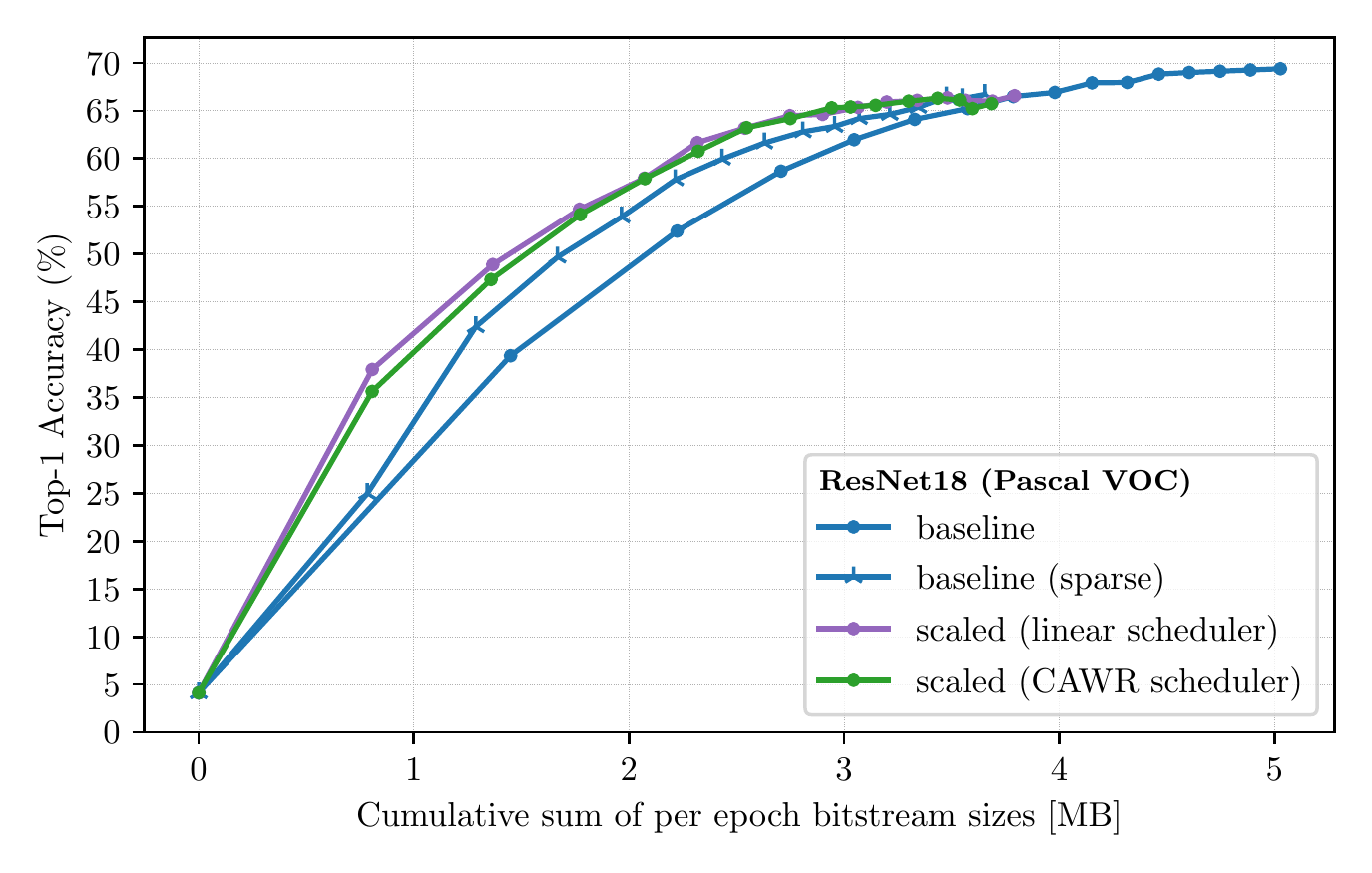}
\includegraphics[width=0.495\textwidth]{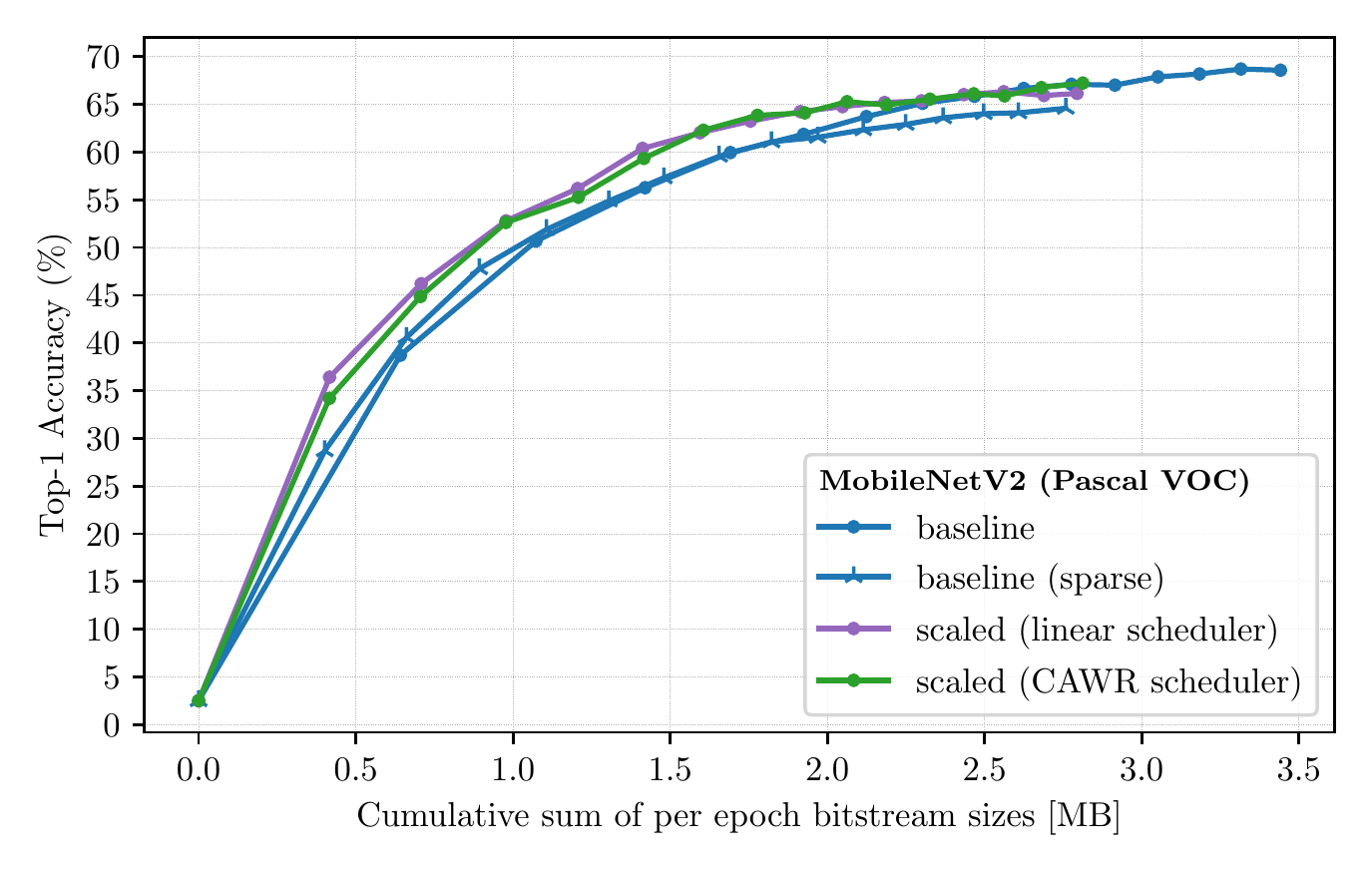}
\includegraphics[width=0.495\textwidth]{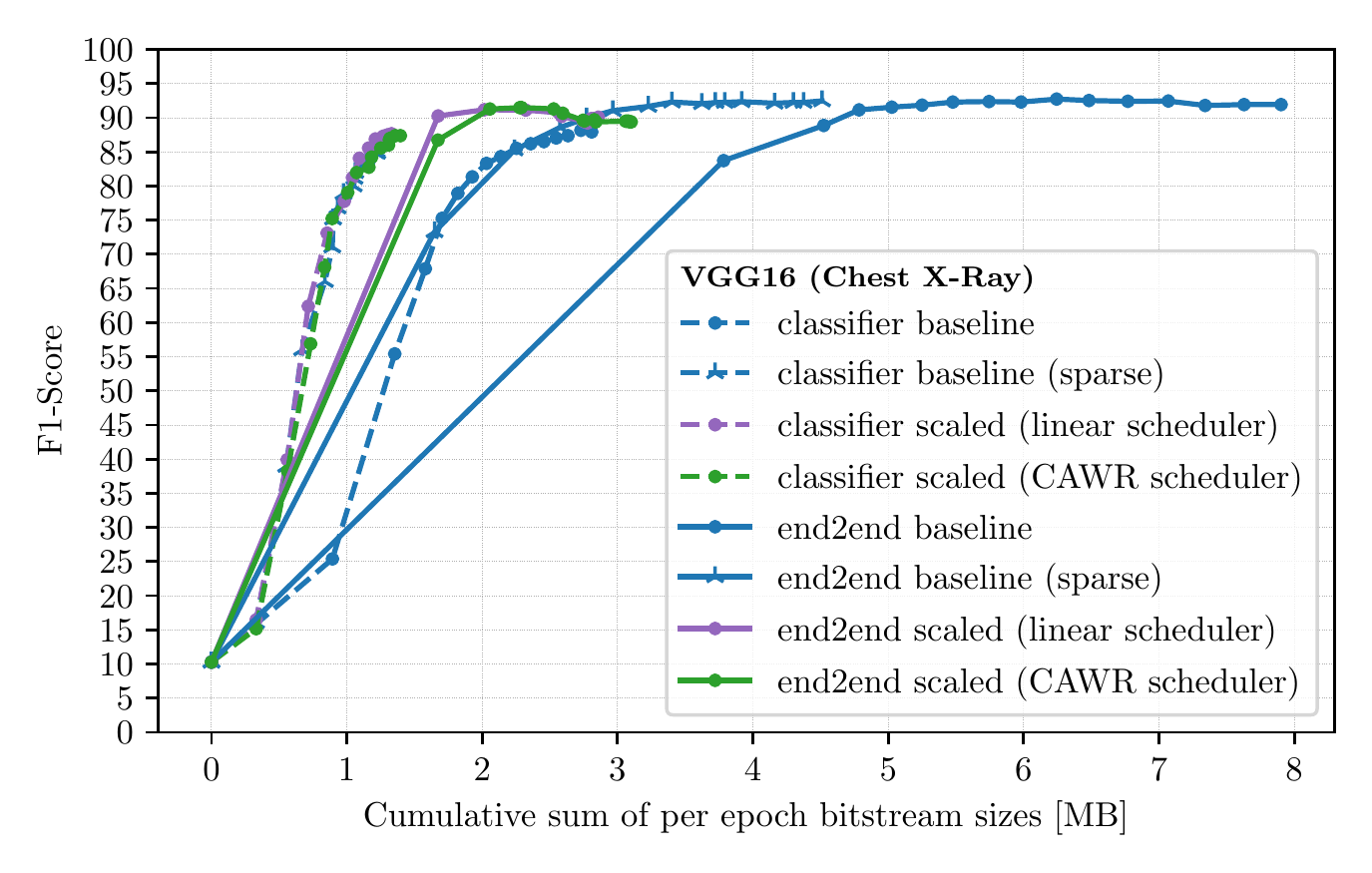}
\caption{SGD optimization for filter-scaled sparse federated training of VGG11, ResNet18, MobileNetV2 and VGG16 (row-major order) solving the Pascal VOC and chest x-ray classification tasks, respectively.}
\label{fig:appendixres}
\end{figure}

\newpage
\section{Data Distribution}
\label{appendix:datadistrib}

\begin{figure*}[ht!]
\centering
\includegraphics[width=0.99\textwidth]{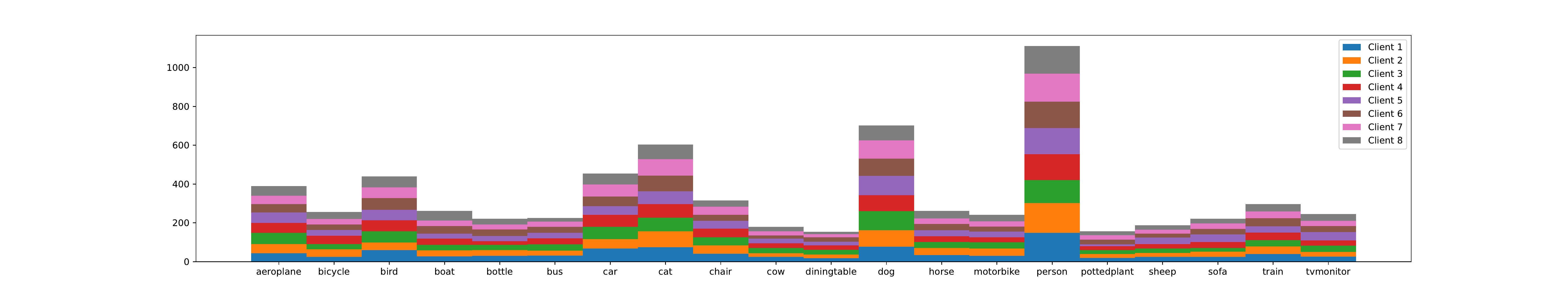}
\includegraphics[width=0.99\textwidth]{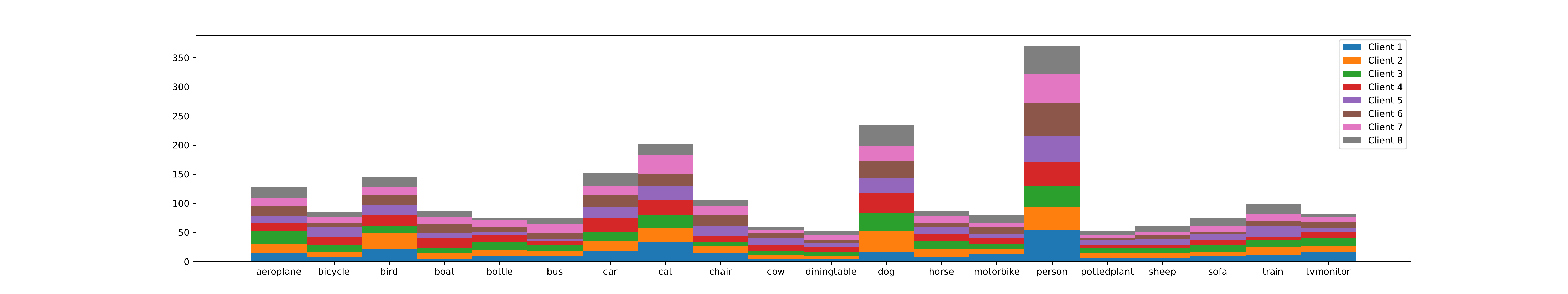}
\caption{Training (top) and validation (bottom) data distribution of the Pascal VOC scenario with 8 clients.}
\label{fig:appendix:datadistrib1}
\end{figure*}

\begin{figure*}[ht!]
\centering
\includegraphics[width=0.99\textwidth]{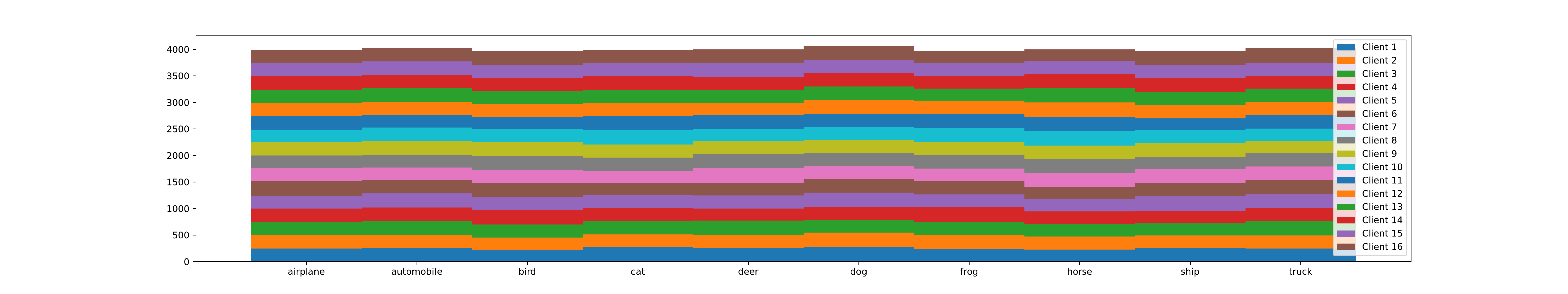}
\includegraphics[width=0.99\textwidth]{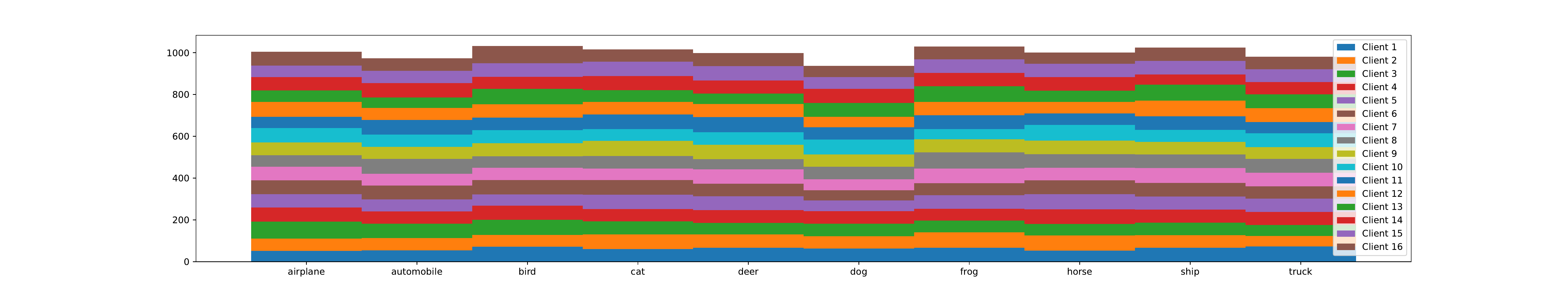}
\caption{Training (top) and validation (bottom) data distribution of the CIFAR10 scenario with 16 clients.}
\label{fig:appendix:datadistrib2}
\end{figure*}

\end{document}